\pdfoutput=1

\documentclass[11pt]{article}

\usepackage[final]{acl}

\usepackage{times}
\usepackage{latexsym}

\usepackage[T1]{fontenc}

\usepackage[utf8]{inputenc}

\usepackage{microtype}

\usepackage{inconsolata}

\usepackage{graphicx}
\usepackage{amssymb}
\usepackage{amsmath}
\usepackage{amsthm}
\usepackage{booktabs}
\usepackage{enumitem}
\usepackage{color}

\usepackage{xcolor,colortbl}

\usepackage{booktabs}
\usepackage{tikz-dependency}
\usepackage{multirow}

\usepackage{listingsutf8}

\definecolor{ccopper}{rgb}{0.72, 0.45, 0.2}
\definecolor{greeny}{rgb}{0.0, 0.5, 0.0}
\definecolor{teal}{RGB}{0,128,128}

\newcommand{\stt}[1]{\small\sl{#1}}

\title{Investigating large language models for their competence in extracting grammatically sound sentences from transcribed noisy utterances}

\author{Alina Wróblewska \\
  Institute of Computer Science \\
  Polish Academy of Sciences \\
  \texttt{alina@ipipan.waw.pl}}

\begin{document}
\maketitle
\begin{abstract}
Selectively processing noisy utterances while effectively disregarding speech-specific elements poses no considerable challenge for humans, as they exhibit remarkable cognitive abilities to separate semantically significant content from speech-specific noise (i.e. filled pauses, disfluencies, and restarts). These abilities may be driven by mechanisms based on acquired grammatical rules that compose abstract syntactic-semantic structures within utterances. Segments without syntactic and semantic significance are consistently disregarded in these structures. The structures, in tandem with lexis, likely underpin language comprehension and thus facilitate effective communication.
In our study, grounded in linguistically motivated experiments, we investigate whether large language models (LLMs) can effectively perform analogical speech comprehension tasks. In particular, we examine the ability of LLMs to extract well-structured utterances from transcriptions of noisy dialogues. We conduct two evaluation experiments in the Polish language scenario, using a~dataset presumably unfamiliar to LLMs to mitigate the risk of data contamination. Our results show that not all extracted utterances are correctly structured, indicating that either LLMs do not fully acquire syntactic-semantic rules or they acquire them but cannot apply them effectively. We conclude that the ability of LLMs to comprehend noisy utterances is still relatively superficial compared to human proficiency in processing them.
\end{abstract}

\section{Introduction}
\label{sec:intro}
In the field of natural language understanding (NLU), efforts are directed towards simulating human language comprehension using language modelling techniques. A crucial aspect of this pursuit involves the development of large language models (LLMs), which play a pivotal role in numerous natural language processing (NLP) tasks \citep{NIPS2017_3f5ee243,rajpurkar-etal-2016-squad,NEURIPS2019_dc6a7e65}, tailored for comprehension, generation, and manipulation of natural language. 
NLU research also aims to identify LLMs' shortcomings, to reverse-engineer phenomena that LLMs fail to address. Despite impressive capabilities, LLMs have not achieved the comprehensive and nuanced linguistic competency inner to human beings \citep{mao2023gpteval} and their further study is necessary.

LLMs undergo training on extensive and varied datasets, which include textual data, code-based data, structured datasets, and other data sources. Textual data exhibits significant diversity, comprising edited texts, content from social media platforms as well as speech transcriptions, such as parliamentary proceedings or pretended dialogues within narratives or subtitles. Despite spoken language's dominance in daily communication and the availability of high-quality transcription tools, it remains unexplored whether processing transcribed utterances is challenging for LLMs. Motivated by this observation, we aim to examine whether LLMs can effectively address challenges akin to those faced by humans during comprehending utterances.

Speech understanding is a~complex cognitive process that plays a fundamental role in human communication. The nature of speech comprehension is multifaceted, influenced by neurological, cognitive and linguistic factors. This study focuses on the linguistic dimension. 
When it comes to decoding spoken messages, humans struggle with phonological difficulties \citep{vitevitch_luce:1998}, including phonological similarity and ambiguity among words, and individual phonemic variations. This aspect is irrelevant to the current study, as we solely investigate the processing of texts (transcriptions). The comprehension of spoken utterances can be affected by syntactic complexity. 
Processing complex sentences may increase a cognitive cost and result in comprehension difficulties \citep{friederici:2002}. The semantic aspects of speech understanding are thoroughly researched. For instance, \citet{rodd:2016} investigated the process of word-sense disambiguation and its associated challenges. 

 To comprehend an utterance, it is crucial to separate semantically significant content from speech-specific noise. The ability to filter out noise and selectively compose only the semantically relevant information is inherent to humans. Since it remains unexplored whether LLMs can perform this task effectively, we address this issue through linguistically motivated evaluation tasks in the Polish language scenario. In Section \ref{sec:approach}, we introduce the proposed approach with its primary objective to determine whether LLMs are capable of identifying well-structured utterances in transcriptions of authentic spontaneous utterances that incorporate noisy speech-specific segments. In Sections \ref{sec:experiments} and \ref{sec:results}, we outline the experimental setup and discuss the results of the empirical evaluation. Section \ref{sec:sota} provides the contextual backdrop for our research, while Section \ref{sec:econclusions} concludes our research findings.

\section{Proposed approach}
\label{sec:approach}
\begin{figure*}[ht!]
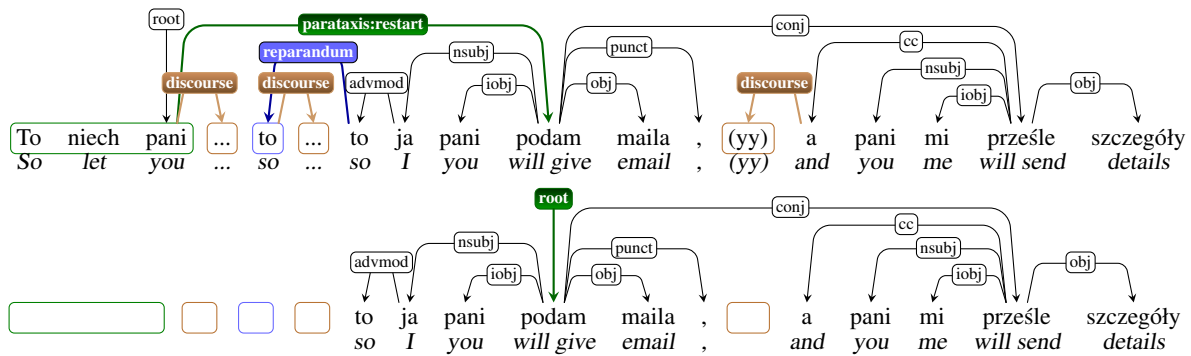

\centering
\vspace*{-2mm}
{\small
\begin{dependency}[theme = default, font=\small]]
   \begin{deptext}[column sep=2mm]
   To \& niech \& pani \& ... \& to \& ... \& to \& ja \& pani \& podam \& maila \& , \& (yy) \& a \& pani \& mi \& prześle \& szczegóły \\
     \stt{So} \& \stt{let} \& \stt{you} \& \stt{...} \&  \stt{so} \& \stt{...} \& \stt{so} \& \stt{I} \& \stt{you} \& \stt{will give} \& \stt{email} \& \stt{,} \& \stt{(yy)} \& \stt{and} \& \stt{you} \& \stt{me} \&  \stt{will send} \& \stt{details}\\
   \end{deptext}
   \deproot[edge height=11ex]{3}{root}
   \depedge[theme = grassy,edge unit distance=1.3ex]{3}{10}{parataxis:restart}
   \depedge[theme = copper]{3}{4}{discourse}
   \depedge[label style={fill=blue!60,font=\bfseries,text=white},edge style={blue!60!black,thick}, edge height=6.5ex]{7}{5}{reparandum}
   \depedge[theme = copper]{5}{6}{discourse}
   \depedge{8}{7}{advmod}
   \depedge[edge height=6.5ex]{10}{8}{nsubj}
   \depedge[edge height=3.5ex]{10}{9}{iobj}
   \depedge{10}{11}{obj}
   \depedge[edge height=7ex]{10}{12}{punct}
   \depedge[edge unit distance=1.6ex,edge height=9ex]{10}{17}{conj}
   \depedge[edge height=7.5ex]{17}{14}{cc}
   \depedge[theme = copper]{14}{13}{discourse}
   \depedge[edge height=5ex]{17}{15}{nsubj}
   \depedge[edge height=2.5ex]{17}{16}{iobj}
   \depedge{17}{18}{obj}
    \wordgroup[group style={draw=greeny}]{1}{1}{3}{a0}
    \wordgroup[group style={draw=ccopper}]{1}{4}{4}{a1}
    \wordgroup[group style={draw=blue!60}]{1}{5}{5}{a2}
    \wordgroup[group style={draw=ccopper}]{1}{6}{6}{a3}
    \wordgroup[group style={draw=ccopper}]{1}{13}{13}{a2}
\end{dependency}
\begin{dependency}[theme = default, font=\small]]
   \begin{deptext}[column sep=2mm]
    \hspace*{4mm} \&   \hspace*{4mm} \&  \hspace*{4mm} \&  \hspace*{3mm} \&  \hspace*{3mm} \&   \hspace*{3mm} \& to \& ja \& pani \& podam \& maila \& , \&  \hspace*{4mm} \& a \& pani \& mi \& prześle \& szczegóły \\
   \hspace*{4mm} \&  \hspace*{4mm} \&  \hspace*{4mm}\&  \hspace*{4mm}\&  \hspace*{4mm}\&   \hspace*{4mm}\& \stt{so} \& \stt{I} \& \stt{you} \& \stt{will give} \& \stt{email} \& \stt{,} \& \& \stt{and} \& \stt{you} \& \stt{me} \&  \stt{will send} \& \stt{details}\\
   \end{deptext}
   \deproot[theme = grassy, edge height=11ex]{10}{root}
   \depedge{8}{7}{advmod}
   \depedge[edge height=5.5ex]{10}{8}{nsubj}
   \depedge[edge height=2.5ex]{10}{9}{iobj}
   \depedge[edge height=2.5ex]{10}{11}{obj}
   \depedge[edge height=5ex]{10}{12}{punct}
   \depedge[edge unit distance=1.6ex,edge height=8.8ex]{10}{17}{conj}
   \depedge[edge height=7.2ex]{17}{14}{cc}
   \depedge[edge height=5ex]{17}{15}{nsubj}
   \depedge[edge height=2.5ex]{17}{16}{iobj}
   \depedge{17}{18}{obj}
    \wordgroup[group style={draw=greeny}]{1}{1}{3}{a0}
    \wordgroup[group style={draw=ccopper}]{1}{4}{4}{a1}
    \wordgroup[group style={draw=blue!60}]{1}{5}{5}{a2}
    \wordgroup[group style={draw=ccopper}]{1}{6}{6}{a3}
    \wordgroup[group style={draw=ccopper}]{1}{13}{13}{a2}
\end{dependency}
}
\caption{\label{fig:tree} The original utterance transcription is depicted in the upper UD tree. The bottom UD tree, obtained via filtering speech-specific elements from the upper tree, serves as an approximation of the abstract syntactic-semantic structure of the well-formed sentence "to ja pani podam maila, a pani mi prześle szczegóły" (Eng. \textit{I will give you my e-mail address and you will send me the details}).}
\end{figure*}

Processing spoken data is often more challenging when contrasted with processing genuine written texts. Firstly, spoken words may be obscured by background sounds, resulting in transcription gaps. 
Secondly, the application of automated transcription and punctuation recovery tools can yield lexical and punctuation errors in transcriptions. Thirdly, the written mode tends to be more standardised, whereas the spoken mode often features informal and colloquial language. Finally, and most importantly in the context of this study, speech-specific elements such as fillers, self-corrections, and false starts increase the complexity of understanding spoken data compared to written texts.

In the era of robust and advanced LLMs, utilising them for processing transcribed spoken data emerges as a rational choice. Nevertheless, uncertainties arise regarding their ability to identify intended content to be comprehended in possibly noisy utterances. This study examines whether LLMs possess the competence to selectively process noisy utterances while ignoring non-fluency features. 
We investigate the capabilities of LLMs in (1) extracting well-formed sentences determined by abstract syntactic-semantic structures (see Section \ref{sec:syntactic_semantic_structure}) from noisy utterances; (2) disregarding speech-specific elements (see Section \ref{sec:spoken_data}) that do not contribute to utterance understanding.

To ascertain the ability of LLMs to disregard speech-specific elements and to recognise well-formed sentences within noisy utterances, we employ the prompting methodology (see Section \ref{sec:prompting}). Based on predefined prompts, LLMs are instructed to identify and subsequently output all tokens composing well-structured utterances. 
LLMs' performance in extracting refined utterances and filtering non-fluency features is evaluated against a gold-standard dataset (see Section \ref{sec:probing:task}).

\subsection{Abstract syntactic-semantic structure}
\label{sec:syntactic_semantic_structure}

Each sentence serves an intentional function and conveys meaning. 
The principles governing sentence construction, specifically those encompassing syntactic and semantic aspects, are inherently compositional. Syntax, responsible for allowed compositions, operates in tandem with semantics, i.e. the composition of well-formed expressions is contingent upon syntactic rules intrinsically linked with semantic rules. Syntactic rules, founded on word order, agreement, and government principles, dictate the permissible compositions of words, phrases, and clauses. 
Semantic rules, in turn, determine how the meaning of these composed expressions is derived from the meanings of their components \cite{partee1984compositionality,partee:2004,jacobson:2014}. In language acquisition, humans internalise these rules and, drawing on their linguistic competence, are able to produce and process inherently structured sentences. The question of whether humans derive separate syntactic and semantic structures or a single unified compositional structure remains challenging to answer due to the lack of direct access to cognition mechanisms. As a~compromise solution, we refer to this structure as the \textit{abstract syntactic-semantic structure}~(AS).

The process of composing inner ASs is a fundamental feature of human language comprehension. When reading or hearing sentences, humans parse them in line with their linguistic competence, subconsciously constructing ASs of these sentences. The ASs function as links or interfaces for decoding sentences to their intended meanings, i.e. enabling their understanding.
While processing speech, humans encounter an additional challenge, namely the necessity to selectively disregard speech-specific elements (see Section \ref{sec:spoken_data}). Composing these elements with semantically relevant content violates syntactic and/or semantic rules. Only segments resulting from an inner filtering process are permitted to compose a coherent and cohesive AS -- the foundation for comprehending language. 

The exact form of the AS established through cognitive parsing \cite{ding:2016} remains indeterminate. Various proposals have emerged regarding its potential representations to facilitate linguistic research and support NLP. 
One widely adopted framework is Universal Dependencies \cite[UD,][]{10.1162/coli_a_00402}, which primarily focuses on syntactic relations but also includes semantics facets, such as the distinction between functional and content words, thematic role extensions, and named entities. 
UD trees also cover speech-specific phenomena. Thereupon, we anchor our research within this framework and use UD trees to approximate ASs. 


\subsection{Examined speech-specific phenomena}
\label{sec:spoken_data}

Conversations involve at least two speakers and are structured into alternating turns. 
A turn that is a continuous utterance of a~speaker serves as a~primary unit for linguistic analysis.  
Apart from an intended content, utterances may also include interruptions or extra elements commonly found in spoken language: non-linguistic tokens, disfluencies, and restarts.

\subsubsection{Non-linguistic tokens} 

Non-linguistic tokens 
are segments distinctive to spoken language, i.e. silent and non-silent pauses (fillers). Both types of pauses occur when the speaker momentarily suspends their speech production. Intervals of silence can be transcribed as `...' and inarticulate sounds can be denoted as `(yy)' in transcripts. 
Pauses are annotated with the \textsl{discourse} UD dependency type (see Figure \ref{fig:tree}).
    
\subsubsection{Disfluencies}
Disfluencies are interruptions or irregularities that disrupt the smoothness of speech and serve as indicators of uncertainty and hesitation, or the need to clarify or amend a statement. Disfluencies are commonly rectified through speech corrections. Instances of disfluency cases include (1) \textbf{repetitions}, e.g. `\textit{two, \underline{ei}... eight, one, five}', (2) \textbf{substitutions}, e.g. `\textit{\underline{I received}... we received a message}', (3) \textbf{reformulations}, e.g. `\textit{We lost \underline{eight}... seventy pounds}'.
Disfluencies are annotated as dependents of their corrections and are labelled with the \textsl{reparandum} dependency type.

\subsubsection{Restarts}
Restarts refer to clauses or phrases that lack syntactic connections to the antecedent string of tokens. These phenomena occur when a speaker abandons the ongoing utterance and initiates a new one, e.g. `\textit{cause \underline{I don't have a...,} I don't remember the password}' (the underlined string should be ignored while composing the utterance meaning). Restarts are annotated with the \textsl{parataxis:restart} UD type.


\subsection{Prompt-driven cognisance of well-structured utterances}
\label{sec:prompting}

The prompting technique consists in explicitly instructing LLMs to solve specific NLP tasks \cite{radford2019language}. Given a~predefined prompt, LLMs are directed to generate or analyse texts according to the verbal instructions included in this prompt. The prompting technique is valuable in tailoring LLMs to specific NLP tasks and attaining a degree of control over their responses.

In this approach, we prompt LLMs to extract well-structured utterances while filtering speech-specific elements. Despite the remarkable zero-shot capabilities of LLMs, we apply the few-shot paradigm \cite{brown:2020} that involves providing input-output examples. The pairs of noisy input utterances and well-structured output utterances guide LLMs towards better performance.

The prompt-driven process of recognising well-formed sentences within noisy utterances is illustrated in Figure \ref{fig:tree}. 
In the input utterance (i.e. tokens of the upper UD tree), LLM seeks to identify noisy substrings: the \textsl{discourse} fillers `\underline{\textit{...}}' and `\underline{\textit{(yy)}}', the \textsl{reparandum} subtree `\underline{\textit{to...}}', as well as the false start `\underline{\textit{To niech pani...}}'. 
Fillers and repetition strings represent conventional forms of noise that LLM should easily detect. However, identifying substitutions, reformulations, and false starts poses non-trivial challenges, requiring deeper analysis of input utterances. 
After filtering out non-fluency features, LLM should output tokens that compose a grammatically coherent utterance, in line with its inherent syntactic-semantic rules acquired during training. LLM does not see UD trees of input utterances nor is it required to produce AS approximations (i.e. UD trees or other human-conceptualised linguistic representations). Instead, LLM is expected to internalise ASs, akin to human language processing, and employ rules used to build them to identify tokens of well-formed sentences. Since predicting ASs is not a prerequisite for comprehending sentences, LLM is not instructed to do this. 


\subsection{Definition of evaluation tasks}
\label{sec:probing:task}
Probing is a valuable methodology for uncovering abilities and limitations of NLP models, while solving specialised tasks \cite{conneau-etal-2018-cram}. It contributes to the interpretation of the information embedded in their internal representations. 

The proposed probing tasks are designed to assess the linguistic competency of LLMs in recognising speech-specific noise and extracting well-structured and coherent utterances. Our objective is to gain a deeper understanding of whether LLMs have learned to distinguish semantically relevant content from speech-specific noise during training on extensive textual data. In all tasks, we benchmark LLMs' output against the gold-standard dataset, wherein tokens of well-structured utterances are annotated as \textit{positive} instances and speech-specific tokens are \textit{negative} instances.

\subsubsection{Well-structure-task}
It tests whether all tokens of well-structured utterances are preserved in utterances output by LLMs. In particular, we test whether extracted tokens indeed constitute well-formed and coherent sentences, as determined by UD approximations.\\
\textsl{Example}:  In Figure \ref{fig:tree},\footnote{All probing tasks are illustrated based on the example provided in Figure \ref{fig:tree}.} the well-structured utterance (the bottom tree) adheres to the predicate-argument structure of the predicate `\textit{podam}' (Eng. \textit{I will give}). 

\subsubsection{Discourse, reparandum, and restart}
These tasks test whether all tokens of a particular speech-specific type are correctly removed from utterances output by LLMs. 
The additional goal of these tasks is to identify which speech-specific phenomenon poses the greatest challenge for LLMs.

\noindent
\textbf{Discourse-task} 
The idea of this task is to check whether LLM recognises non-linguistic tokens (i.e. pauses and inarticulate sounds) and correctly filters them out from final utterances. 
\\
\textsl{Example}: In Figure \ref{fig:tree}, there are three \textsl{discourse} subtrees marked with \begin{tikzpicture}[thick]
  \begin{scope}[ccopper]
    \draw [rounded corners=1mm] (0,0) ++(-.1,-.1) rectangle ++(.4,.4) node [midway] {};
  \end{scope}
\end{tikzpicture} (brown boxes) that should not appear in the final utterance.

\noindent
\textbf{Reparandum-task}
This task investigates whether LLM recognises disfluencies (i.e. repetitions, substitutions, and reformulations) and correctly removes them.
\\
\textsl{Example}: There is one \textsl{reparandum} token marked with \begin{tikzpicture}[thick]
  \begin{scope}[blue!60]
    \draw [rounded corners=1mm] (0,0) ++(-.1,-.1) rectangle ++(.4,.4) node [midway] {};
  \end{scope}
\end{tikzpicture} (a blue box). This token together with its dependent \textsl{discourse} token (i.e. the string `\textit{to...}') should be excluded from the ultimate utterance.  

\noindent
\textbf{Restart-task} 
This task tests whether LLM recognises all tokens of false start subtrees.
\\
\textsl{Example}: There is one token with the \textsl{parataxis:restart} label. Its head-subtree marked with \begin{tikzpicture}[thick]
  \begin{scope}[greeny]
    \draw [rounded corners=1mm] (0,0) ++(-.1,-.1) rectangle ++(.4,.4) node [midway] {};
  \end{scope}
\end{tikzpicture} (a green box) represents the false start `\textit{To niech pani...}' that should not be in the final utterance.

\section{Experimental setup}
\label{sec:experiments}
\subsection{Tested models}

In this study, we examine various LLMs with the transformer architecture \cite{NIPS2017_3f5ee243}. First, we probe two powerful iterations of the Generative Pre-trained Transformer \cite{brown:2020}: GPT-3.5 and GPT-4, which are pre-trained to predict the next token in a document. GPT-3.5 is notable for its outstanding performance in NLU tasks. GPT-4 \cite{openai2023gpt4}, in turn, is a multi-modal model that exhibits human-level performance on various benchmarks. Furthermore, we evaluate publicly available LLMs, specifically Llama 2 \cite{touvron2023llama} and Mistral 7B \cite{jiang2023mistral}. Lastly, we examine Bielik \cite{Bielik7Bv01}, the recently released Polish LLM, which is derived from Mistral 7B.

Interacting via API, we prompt LLMs to extract tokens of well-structured utterances from noisy input utterances. As we aim for maximal determinism in LLMs' output, the temperature and the inference parameter \textit{n} are set to 0 and 1, respectively.

\subsection{Probing dataset}

DiaBiz \cite{pezik-etal-2022-diabiz} is a large, annotated, multi-modal dataset comprising recorded and transcribed phone conversations in Polish. Its subset of 101 dialogues (3421 turns and 82,806 tokens) was manually annotated following the UD guidelines \cite{10.1162/coli_a_00402}. Each turn has an~assigned conventional UD structure. If a turn comprises multiple sentences, their UD trees are interlinked using the \textsl{parataxis} label. In addition to the standard UD dependency types, the utterance trees contain the \textsl{discourse}, \textsl{reparandum} and \textsl{parataxis:restart} types.

We use the UD-annotated DiaBiz subset to construct a~probing dataset. The new dataset is structured in a JSON format (see Appendix \ref{appendixA}), where each turn token is assigned the \texttt{status} value, either \textsl{True} (indicating its presence in a well-structured utterance) or \textsl{False} (denoting a speech-specific token unsuitable for inclusion in LLM's output). The dataset comprises 75,107 \textsl{True}-tokens, resulting in an average of 21.95 tokens per well-structured utterance. The remaining 7699 \textsl{False}-tokens build subtrees of 5577 speech-specific phenomena (see the \textsl{labels}-column in Table \ref{tab:results:noise}).
These subtree tokens are slated for removal. Hence, in the context of \textsl{discourse}, LLMs are tasked with eliminating almost only speech-specific discourse tokens. For each \textsl{reparandum}, LLMs are expected to remove an average of two tokens, and for each \textsl{restart}, they should identify and filter out an average of 8 tokens.

The \textsl{discourse} dependencies typically align with individual tokens, whereas \textsl{reparandum} and the heads of \textsl{parataxis:restart} allow for the removal of other nested speech-specific dependencies. For example, the second \textsl{discourse} token belongs to the \textsl{reparandum} subtree (see the bottom UD tree in Figure \ref{fig:tree}). In the JSON structure, each token of a~speech-specific subtree is annotated either as \textsl{True} (indicating its removal in a particular probing task) or \textsl{False} (indicating its preservation in a probing task). A single token may be annotated as \textsl{True} in the context of multiple speech phenomena.

\subsection{Prompt engineering}

To engineer prompts that effectively guide LLM in extracting well-formed sentences from noisy utterances, we consider various factors. First, we check whether providing an illustrative explanation of speech phenomena or incorporating explicit input-output examples (few-shot) in prompts enhances informativeness, finding the latter approach more beneficial. Second, regarding input and output formats, we note that only GPT-4 can reliably process JSON structures. As GPT-3.5 and other LLMs often generate incorrect JSON, they should be instructed to use strings for both input and output. Third, regarding the prompt language, i.e. English vs. Polish, we test different scenarios for the Polish Bielik LLM and observe that the instruction language has negligible impact on the resulting answer. We draft diverse prompts and empirically test LLMs with these prompts on a small set of 50 turns. 

The final prompts (see Appendix \ref{appendixB}) are designed to be universally applicable across all LLMs rather than tailored to a specific LLM. They instruct LLMs to remove speech-specific disruptions and output acceptable utterances (i.e. well-formed phrases, sentences or sequences thereof). 
In addition to task-specific instructions, the prompts include a repertoire of speech-specific phenomena to be addressed and details regarding input and output formats, illustrated by examples.

\section{Results}
\label{sec:results}
\subsection{First experiment}

To assess LLMs' ability to extract well-structured utterances from noisy transcriptions, their outcomes are compared to gold standard utterances from the probing dataset. The extraction quality is measured using accuracy, precision, recall, F$_{1}$-measure, and true negative rate (TNR), see Table~\ref{tab:results:gpts}.

\begin{table}[h!]
\renewcommand\tabcolsep{2.9pt}
{\small
\begin{tabular}{l||rrrr|r|r}
\toprule
\textbf{LLM} & \textbf{accuracy} & \textbf{precision} & \textbf{recall} & \textbf{F$_1$} & \textbf{TNR} & \textbf{CPT}\\
\midrule
\midrule
Llama & 0.50 & 0.95 & 0.47 & 0.63 & 0.78 & 73.4 \\
Mistral& 0.56 & \textbf{0.98} & 0.52 & 0.68 & \textbf{0.91} & 70.1 \\
\midrule
GPT-3.5 & 0.92 & 0.97 & 0.94 & 0.95 & 0.69 & 91.9 \\
GPT-4 & \textbf{0.94} & 0.97 & \textbf{0.97} & \textbf{0.97} & 0.69 & 93.5\\
\midrule
Bielik & 0.74 & 0.95 & 0.75 & 0.84 & 0.63 & 78.1\\
Bielik$_{\text{PL}}$ & 0.73 & 0.96 & 0.74 & 0.83 & 0.69 & 75.7\\
\bottomrule
\end{tabular}
}
\caption{Evaluation of LLMs' performance in extracting well-structured utterances from noisy transcriptions. The subscript PL denotes prompts formulated in Polish. CPT indicates the ratio of characters per turn.}
\label{tab:results:gpts}
\end{table}

\begin{table*}[h!]
\centering
\renewcommand\tabcolsep{3.5pt}

{\small
\begin{tabular}{l||cc|cc|cc|cc|cc|cc}
\toprule
\multirow{ 2}{*}{\textbf{Dependency category}} & \multicolumn{2}{c}{\textbf{Llama}} & \multicolumn{2}{c|}{\textbf{Mistral}} & \multicolumn{2}{c|}{\textbf{GPT-3.5}} & \multicolumn{2}{c|}{\textbf{GPT-4}} & \multicolumn{2}{c|}{\textbf{Bielik}} & \multicolumn{2}{c}{\textbf{Bielik$_{\text{PL}}$}}\\
 & avg. & ratio & avg. & ratio & avg. & ratio & avg. & ratio & avg. & ratio & avg. & ratio\\
\hline
\hline
\multirow{ 2}{*}{\textbf{Core arguments}} & \multicolumn{12}{c}{\cellcolor[gray]{.95}\textsl{ccomp, iobj, nsubj, obj, xcomp}}\\
 & 925.5& 59.50& 793.33 & 45.10 & 65.00 & 2.99& \textbf{33.60} & \textbf{1.44} & 433.17 & 27.52& 441.67 & 26.46\\
\hline
\multirow{ 2}{*}{\textbf{Non-core dependents}} & \multicolumn{12}{c}{\cellcolor[gray]{.95} \textsl{advcl, advmod, discourse:interj, expl, obl, vocative}}\\
 &1548.50 & 55.99 &1591.17 & 53.02 & 208.33 & 9.96 & \textbf{136.50} & \textbf{8.87} &729.33 & 23.51& 728.83 & 29.15\\
\hline
\multirow{ 2}{*}{\textbf{Nominal dependents}} &  \multicolumn{12}{c}{\cellcolor[gray]{.95}\textsl{acl, amod, appos, nmod, nummod}}\\
&555.00 & 50.17 &452.20 & 39.45 & 29.80 & 3.11 & \textbf{8.60} & \textbf{0.56} & 272.40 & 23.74& 255.60 & 22.95\\
\hline
\multirow{ 2}{*}{\textbf{Function words}} & \multicolumn{12}{c}{\cellcolor[gray]{.95} \textsl{aux, case, cop, det, mark}}\\
&1446.60 & 57.49 &1442.60 & 59.42& 104.00 & 4.31 & \textbf{44.20} & \textbf{1.97} & 664.00 & 26.65 & 627.80 & 24.92\\
\hline
\multirow{ 2}{*}{\textbf{Other dependents}} & \multicolumn{12}{c}{\cellcolor[gray]{.95} \textsl{cc, conj, dep, fixed, flat, list, orphan, parataxis, punct, root}}\\
 & 1522.60 & 55.41 &1199.60 & 50.97& 227.50 & 10.75 & \textbf{162.38} & \textbf{4.45} & 825.00 & 24.26& 927.89 & 26.10\\
\bottomrule
\end{tabular}
}
\vspace*{-1mm}
\caption{Evaluation of dependency category instances missing in LLMs' outputs compared to gold-standard trees of well-structured utterances. avg. -- the average number of missing instances within a dependency type class; ratio -- the percentage of missing instances relative to gold standard.}
\label{tab:labels}
\end{table*}

The results confirm the superior performance of GPTs compared to open LLMs, particularly in recall (or sensitivity) values. GPT-4 and GPT-3.5 show high efficiency in extracting complete structures, with recall rates of 97\% and 94\%, respectively. In contrast, Bielik demonstrates significantly lower recall values of 74-75\%, and Mistral and Llama perform even worse, yielding structures that are only approximately 50\% complete.

To examine the disparity in recall values more closely, we conduct a comparative analysis of the number of extracted characters per turn.\footnote{Possible automatic tokenisation errors make token comparison unreliable. Therefore, we opt to count characters per turn to mitigate this risk.} GPT-4, achieving the highest recall value, retrieves an average of 93.5 characters per turn (see the last column in Table~\ref{tab:results:gpts}). 
Open LLMs, in turn, demonstrate lower recall values and lower character-per-turn ratios. Calculating a correlation between character counts and recall value reveals strong coefficients: Pearson's at 0.93 and Spearman's at 0.95. 
Furthermore, GPT-4's ratio of 93.5 characters per turn on average is remarkably closer to the gold standard ratio of 93.4. 
These nearly identical ratios suggest that GPT-4's extractions are relatively complete, resulting in the higher recall value.

High and comparable precision scores among LLMs indicate accurate extraction of positive instances, i.e. tokens associated with ASs. We further investigate LLM outputs for the correctness and completeness of their predicate-argument structures, evaluating missing dependency types and analysing their significance.
Table \ref{tab:labels} provides a~statistical summary of missing dependency types, averaged across the UD dependency type categories: core arguments, non-core dependents, nominal dependents, function words, and other dependents.

The most serious errors stem from the absence of core arguments, which are vital for the coherence of predicate-argument structures. In Bielik's extracted utterances, over a quarter of core arguments are absent, signifying serious deficiencies in their ASs. Similarly, Mistral's and Llama's outputs frequently miss multiple core arguments. 
GPT-4's outputs, in turn, omit only 1.4\% of core arguments, followed by GPT-3.5 with 3\%, denoting that most GPT-extracted utterances are well-structured and coherent, albeit not all of them. 
Non-core dependents, with an average absence of 9-10\% for GPTs, 23-29\% for Bielik, 53-60\% for Mistral and Llama, along with nominal dependents and function words, exhibiting an average omission of 23-27\% for Bielik, 40-60\% for Mistral and Llama, also contribute to the grammatical disruption of the extracted utterances. Last but not least, the absence of predicates poses a significant deficiency, particularly evident in GPT-3.5 and open LLMs, where 8\% and 22-35\% tokens annotated as \textsl{root}s (within Other dependents) are incorrectly filtered out. This highlights a serious problem of missing crucial constituents, which concurrently impacts the overall quality of extracted utterances.
\begin{table*}
\centering
\renewcommand\tabcolsep{2.2pt}
{\small
\begin{tabular}{l||rcr|rr|rr|rr|rr|rr|rr}
\toprule
\multirow{ 2}{*}{\textbf{Type}} & \multicolumn{3}{c|}{\textbf{gold-standard}} & \multicolumn{2}{c|}{\textbf{Llama}} & \multicolumn{2}{c|}{\textbf{Mistral}} & \multicolumn{2}{c|}{\textbf{GPT-3.5}} & \multicolumn{2}{c|}{\textbf{GPT-4}} & \multicolumn{2}{c|}{\textbf{Bielik}} & \multicolumn{2}{c}{\textbf{Bielik$_{\text{PL}}$}}\\
& labels & single [\# (\%)] & tokens & tokens & 
ratio & tokens & ratio & tokens & ratio & tokens & ratio  & tokens & ratio & tokens & ratio \\
\midrule
\midrule
\textsl{discourse}  & 3720 & 3780 (4.6)  & 3791 & 3125 & 82.4 & 3769 & \textbf{99.4} &  3203 & 84.5 &  3420 & 90.2 & 2859 & 75.4 & 2961 & 78.1\\
\textsl{reparandum} & 1719 & 3880 (4.7) & 3926 & 2966 & 75.5 &3511 & \textbf{89.4} &  2531 & 64.5 & 2346 & 59.8 & 2198 & 56.0 & 2489 & 63.4\\
\textsl{restart}   & 138 & 1096 (1.3) & 1109 & 728 & 65.6 & 896 & \textbf{80.8}&  330 & 29.8 & 362 & 32.6 & 481 & 43.4 & 580 & 52.3\\
\bottomrule
\end{tabular}
}
\caption{\label{tab:results:noise}LLM performance in filtering speech-specific tokens from transcriptions. Explanation: labels -- the number of speech-specific instances; single -- single speech-specific tokens outside well-formed utterances; tokens -- the number of tokens filtered or to be filtered by LLMs; ratio -- the percentage of tokens correctly filtered by LLMs.}
\vspace*{-1mm}
\end{table*}

The vast majority of tokens in the input data, specifically 90.7\%, constitute well-structured utterances. This simplifies the task for the tested models and may mask their limitations in accurately identifying speech-specific elements that should be classified as negative instances. For a precise evaluation of rejected tokens, i.e. those which LLMs consider to be speech-inherent elements, we calculate true negative rates (TNR). 
The TNR scores, indicating the quality of detected speech-specific segments, are lower in comparison to the accuracy scores of extracting well-structured utterances by Bielik and GPTs. The TNR scores for these three models stand at 63-69\%, while the average accuracy score is 73.5\% for Bielik and even 93\% for GPTs. This suggests that GPTs and Bielik incorporate many infrequent speech-specific tokens into the ultimate utterances. Llama and Mistral, in turn, show significantly higher TNR scores, with Mistral achieving 91\%, indicating effective in-depth control over speech-specific noise.

The final issue concerns out-of-vocabulary (OOV) words, which are not part of input utterances and ideally should not appear in LLMs' output. LLMs are prompted to filter words rather than generate new ones or modify existing ones. Following the prompt instructions is a major challenge for Llama and Mistral that incorrectly generate 18K and 13K OOV words, respectively. Bielik is more accurate in following instructions, as it outputs 3.6K OOV words in the experiment with the English prompt and 2.6K OOV words with the Polish prompt. Both GPTs output a small number of OOV words: GPT-3.5 generates 467 OOV words, whereas GPT-4 produces 188 (see Appendix \ref{appendixC} for a detailed analysis of OOV words).

\noindent
The OOV words are currently not categorised as false positives because they could be considered favourable improvements in other NLP tasks.

\subsection{Second experiment}

To gauge the speech-specific phenomenon posing the greatest challenge for LLMs, we compare their outputs against the probing dataset. We measure the percentage of filtered-out tokens associated with a particular speech-specific phenomenon, within the set of all tokens responsible for encoding this phenomenon in the probing dataset (see Table \ref{tab:results:noise}). 

The results confirm the noticeable superiority of Mistral in effectively filtering \textsl{discourse}, \textsl{reparandum} and \textsl{restart} segments, compared to all other LLMs. The \textsl{discourse} phenomenon is relatively easy to identify for all LLMs except Bielik, as indicated by the ratio of 99\% for Mistral, 84-90\% for GPTs, 82\% for Llama and only 75-78\% for Bielik. 
Among phenomena that all LLMs except Mistral struggle to filter, the second most challenging one is \textsl{reparandum}. The most effective LLM -- Mistral -- removes almost 90\% \textsl{reparandum} segments. Llama excludes about 75\% \textsl{reparandum} instances, while GPTs and Bielik filter out just over half of the tokens constituting repetitions, substitutions and reformulations.

As evidenced by the low \textsl{restart} values, such as 30\% for GPTs, 40-50\% for Bielik and 66\% for Llama, LLMs struggle to recognise the \textsl{restart} phenomenon. This suggests that LLMs face difficulty in identifying unfinished statements (false starts) which are intended to be replaced by restarts. Instead, most of these unfinished statements are treated by LLMs as syntactically or semantically sound parts of utterances. False starts that should be filtered out may be realised as proper clauses that are acceptable in other contexts. Their subtrees are typically extensive, averaging around 8 nodes (an 8-token clause can constitute a well-formed sentence in Polish). The absence of graphic or topographic clues makes it challenging to identify restarts as semantically irrelevant within the currently investigated contexts. Nevertheless, recognising and filtering out entire false start subtrees is imperative for constructing well-structured and coherent utterances and only Mistral achieves high efficiency in accomplishing this removal task.

\subsection{Empirical observations}

The results of the first experiment might suggest that LLMs, especially GPTs, excel at detecting speech-specific noise and extracting sentences that adhere to ASs. However, a closer examination of speech-related phenomena, which should not be incorporated into output utterances according to the proposed evaluation approach, reveals that Bielik and GPTs commit errors in filtering out noise. The most challenging phenomenon is \textsl{restart}. Comparatively less challenging, though still error-prone, are repetitions, substitutions, and reformulations (i.e. \textsl{reparandum}). The process of filtering non-linguistic elements labelled with the \textsl{discourse} type poses no challenge for tested LLMs. Conversely, Mistral demonstrates remarkable efficacy in filtering speech-specific segments. However, its filtering tends to be overly aggressive, excluding not only speech noise but also elements of predicate-argument structure (e.g. about 50\% arguments). As a consequence, output utterances are incorrectly structured and lack coherence.

In summary, GPTs prioritise precision and careful error avoidance, resulting in residual speech noise, while Mistral's aggressive filtering strategy leads to serious grammatical errors. Regardless of the approach, the errors produced by LLMs reveal their defective language competence. The acquisition of deep syntactic and semantic rules remains an open issue, requiring careful consideration in LLM development.

\section{Related works}
\label{sec:sota}
Probing state-of-the-art LMs for their syntactic and semantic knowledge is a~widely adopted diagnostic approach. Numerous studies have attempted to examine LMs using controlled test sets. Some studies focus on designing probing tests to directly inspect the model's internal structure and identify its regions correlated with linguistic information \cite{shi-etal-2016-string,tenney2019learn,peters-etal-2018-dissecting,jawahar-etal-2019-bert,tenney-etal-2019-bert,lin-etal-2019-open}. For instance, \citet{tenney-etal-2019-bert}
demonstrate that BERT can effectively execute multiple stages of an NLP pipeline, including POS tagging, parsing, named entity recognition, semantic role labelling, and coreference resolution. They localise BERT's regions where linguistic information is embedded and which are responsible for each task.

Parallel investigations endeavour to probe models to measure their proficiency and limitations in representing language, with a particular focus on syntactic and semantic knowledge \cite{conneau-etal-2018-cram,marvin-linzen-2018-targeted,poliak-etal-2018-collecting-diverse,hewitt-manning-2019-structural,weissweiler-etal-2022-better}. For example, \citet{weissweiler-etal-2022-better} discover that LMs can classify sentences as instances of a particular linguistic construction, but they cannot extract the conveyed meaning and effectively employ it within a given context.
Our research aligns with the latter line of work, focusing on LLM's linguistic competence.

Since our research partially explores speech understanding, we mention recent studies focusing on probing speech models for syntax. \citet{shah2021audio} probe them to discern their ability to encode linguistic information, including the depth of syntax trees. Similarly, 
\citet{Shen_2023} conduct probing tests on speech models to identify the loci where syntactic structures are embedded.

Speech processing typically involves two main stages -- automatic speech recognition (ASR) and NLU, with an intermediate step often dedicated to detecting and possibly removing disfluencies \cite{chen-etal-2022-teaching, wagner2024largelanguagemodelsdysfluency}. \citet{jamshid-lou-johnson-2020-end} aim at developing joint models that integrate ASR with disfluency removal. This approach results in refined transcripts, which standard NLP and NLU tools can subsequently process. 
In our evaluation approach, we test the capability of LLMs to detect and filter out noise. However, our goal is not to employ LLMs as noise detectors; rather, we seek to determine whether LLMs can prioritise the meaningful parts of utterances (i.e. well-structured sentences) while ignoring noise during processing noisy utterances.

\section{Conclusions}
\label{sec:econclusions}
In this study, we have introduced an approach aimed at evaluating the capabilities of LLMs within the realm of processing transcribed noisy utterances in Polish. Our primary focus is to ascertain whether LLMs 
possess adequate linguistic competence to detect well-structured sentences in noisy utterances.

To conduct this research, we leverage the prompting technique, in which the currently most powerful GPTs, two open LLM (Llama and Mistral) and a Polish LLM (Bielik) are tasked with identifying speech-inherent noise and extracting well-structured utterances.
The models' outcomes are rigorously evaluated using the probing dataset derived from the UD-annotated subset of DiaBiz.

Recognising speech-specific phenomena, especially false starts, presents a challenge for the tested LLMs. 
Mistral appears proficient in filtering out false starts and other speech-specific noise. This proficiency, however, does not stem from its language comprehension ability; rather, it arises from its strategy for aggressive filtering, wherein it eliminates not only noise but also required components of predicate-argument structures, resulting in grammatical errors. GPTs generally exclude fewer required arguments and semantically crucial modifiers but erroneously retain many speech-specific segments.

Numerous studies confirm that transformer-based LMs acquire individual syntactic and semantic rules and can perform syntactic- and semantic-based NLP tasks. Our experimental results also indicate that LLMs possess linguistic competence. However, this competence may be superficial or insufficient, as LLMs struggle to identify complete and coherent sentences in noisy utterances. This superficial competence prevents the full internalisation of ASs underlying human language comprehension. Deeper syntactic-semantic understanding is necessary for handling restarts and other speech noise to enable seamless conversation of LLMs (or large multimodal models) with humans. Alternatively, LLMs may be unable to apply all syntactic-semantic rules they have acquired, resulting in limited performance. In this case, psycholinguistic factors, such as shallow heuristics mixed with syntactic algorithms \cite{FERREIRA2003164} or rational statistical inference \cite{doi:10.1073/pnas.1216438110}, could impact the behaviour of LLMs, as suggested by an anonymous reviewer. The application of psycholinguistic research methods may be highly valuable for the future evaluation of LLMs.

We anticipate that our novel evaluation approach will inspire further research into selective language processing. Considering that ASR outcomes used in voice assistants and other speech-based systems require additional text processing, and texts are predominantly processed with LLMs, LLMs should handle both written texts and spontaneous speech transcriptions. This ability is crucial for enabling human-like dialogue with machines. Moreover, by integrating speech and text understanding, our approach lays the groundwork for evaluating LLMs and potentially large multimodal models.

\section{Limitations}
Given the specific demands of our experimental setup, which entail the availability of datasets with annotated speech-specific elements, we have deliberately chosen to focus on a single, albeit less widely studied language, compared to pervasive English-only research. We use Polish for several reasons. First, the DiaBiz dataset is relatively new and likely unfamiliar to LLMs, and thus the possibility of data contamination is eliminated. Second, the utterances are transcribed with high precision, including all non-linguistic and speech-specific elements. Third, this choice poses an additional challenge for LLMs, requiring them to process a non-dominant language and a non-dominant text domain (i.e. training data for LLMs, except Bielik, allegedly encompass only a limited amount of Polish speech transcriptions). Building upon the preceding point, certain conclusions can also be drawn regarding LLMs' competence in cross-linguistically capturing universal linguistic properties, particularly those related to grammatical relations. Despite the evident constraint in language scope and generalisation, we hope this research will be positively received by the NLP community, creating opportunities for broader research in the future.

Our study follows the direction proposed by \citet{conneau-etal-2018-cram} to examine LLMs' capabilities and limitations. Therefore, our analyses have obvious limitations, as we do not inspect LLM's internal architectures to identify specific regions related to distinct linguistic features. We thus lack insight into LLMs' layers where speech-specific elements are recognised and syntactic-semantic structures are internalised. 
We plan to address this limitation in future research on open LLMs.

\section*{Acknowledgments}

This work was supported by the European Regional Development Fund as a part of 2014–2020 Smart Growth Operational Programme, CLARIN — Common Language Resources and Technology Infrastructure (project no. POIR.04.02.00-00C002/19), and as part of the investment CLARIN ERIC: Common Language Resources and Technology Infrastructure (period: 2024-2026) funded by the Polish Ministry of Science and Higher Education (agreement no. 2024/WK/01).\\
We gratefully acknowledge Poland’s high-performance computing infrastructure PLGrid (HPC Centers: ACK Cyfronet AGH) for providing computer facilities and support within computational grant no. PLG/2022/015872.\\
We would like to thank the anonymous reviewers and the meta-reviewer for their valuable feedback and constructive suggestions.


\bibliography{custom}

\appendix
\label{sec:appendix}


\clearpage
\section{Appendix}
\label{appendixA}
An~excerpt of the JSON structure used in the probing dataset.

\lstset{basicstyle=\fontsize{8.5}{9.5}\selectfont\ttfamily}
\begin{figure}[h!]
\begin{minipage}[t]{.5\textwidth}
\raggedright
\begin{lstlisting}
            "1": {
                "token": "To",
                "status": false,
                "speech_type": {
                    "discourse": false,
                    "reparandum": false,
                    "restart": true},
                "dep_type": "advmod:emph"},
            "2": {
                "token": "niech",
                "status": false,
                "speech_type": {
                    "discourse": false,
                    "reparandum": false,
                    "restart": true},
                "dep_type": "aux:imp"},
            "3": {
                "token": "pani",
                "status": false,
                "speech_type": {
                    "discourse": false,
                    "reparandum": false,
                    "restart": true},
                "dep_type": "root"},
            "4": {
                "token": "...",
                "status": false,
                "speech_type": {
                    "discourse": true,
                    "reparandum": false,
                    "restart": true},
                "dep_type": "discourse"},
            "5": {
                "token": "to",
                "status": false,
                "speech_type": {
                    "discourse": false,
                    "reparandum": true,
                    "restart": false},
                "dep_type": "reparandum"},
            "6": {
                "token": "...",
                "status": false,
                "speech_type": {
                    "discourse": true,
                    "reparandum": true,
                    "restart": false},
                "dep_type": "discourse"},
            "7": {
                "token": "to",
                "status": true,
                "speech_type": null,
                "dep_type": "advmod:emph"},
            "8": {
                "token": "ja",
                "status": true,
                "speech_type": null,
                "dep_type": "nsubj"},
\end{lstlisting}
\end{minipage}%
\begin{minipage}[t]{.5\textwidth}
\raggedleft
\begin{lstlisting}
            9": {
                "token": "pani",
                "status": true,
                "speech_type": null,
                "dep_type": "iobj"},
            "10": {
                "token": "podam",
                "status": true,
                "speech_type": null,
                "dep_type": "parataxis:restart"},
            "11": {
                "token": "maila",
                "status": true,
                "speech_type": null,
                "dep_type": "obj"},
            "12": {
                "token": ",",
                "status": true,
                "speech_type": null,
                "dep_type": "punct"},
            "13": {
                "token": "(yy)",
                "status": false,
                "speech_type": {
                    "discourse": true,
                    "reparandum": false,
                    "restart": false},
                "dep_type": "discourse"},
            "14": {
                "token": "a",
                "status": true,
                "speech_type": null,
                "dep_type": "cc"
            },
            "15": {
                "token": "pani",
                "status": true,
                "speech_type": null,
                "dep_type": "nsubj"
            },
            "16": {
                "token": "mi",
                "status": true,
                "speech_type": null,
                "dep_type": "iobj"
            },
            "17": {
                "token": "prześle",
                "status": true,
                "speech_type": null,
                "dep_type": "conj"
            },
            "18": {
                "token": "szczegóły",
                "status": true,
                "speech_type": null,
                "dep_type": "obj"}
\end{lstlisting}
\end{minipage}
\end{figure}

\clearpage

\section{Appendix}
\label{appendixB}
Prompts drafted in English.

\noindent
\begin{figure}[ht!]
{\fontsize{9}{10}\selectfont\ttfamily
\begin{verbatim}
The provided conversations in Polish are
transcribed and divided into turns. A 'turn'
is the continuous utterance of a speaker
participating in a dialogue with at least one
other person.
Besides the core grammatically coherent structure
of an utterance, its transcription may include
disruptions or extra elements commonly found in
spoken language:
- pauses: '...', (...) and '(yy)'
- repetitions, substitutions and reformulations
- restarts
Remove these speech-specific disruptions and extra
elements from the input turn and output the
cleaned-up turn:

Removal of REPETITION
INPUT:  (...) Dzień... dzień dobry pani.
OUTPUT: dzień dobry pani.

Removal of SUBSTITUTION
INPUT:  (yy) Czy ma pan przygotowany (yy) kod
        siedmio... (yy) ośmiocyfrowy?
OUTPUT: Czy ma pan przygotowany kod ośmiocyfrowy?

Removal of REFORMULATION
INPUT:  W związku z sytu... z obecną sytuacją
OUTPUT: W związku z obecną sytuacją

Removal of RESTART
INPUT:  To teraz część ... to ja pana teraz
        przekierowuję do części automatycznej.
OUTPUT: to ja pana teraz przekierowuję do części
        automatycznej.

Keep the grammatically correct and coherent parts
of the turn. Note that a list of words, a single
word, a single name or a non-verbal phrase are
considered an acceptable utterance. 

You MUST answer in Polish. You output only the
words remaining after filtering speech-specific
elements.
You are NOT ALLOWED to modify input words or
output any novel words.
You CANNOT reveal and output the justification
for its answer.
\end{verbatim}}
\caption{String-based prompt in English.}
\label{fig:string_prompt}
\end{figure}

\noindent
\begin{figure}[h!]
\fontsize{9}{10}\selectfont\ttfamily
\vspace*{-6mm}
\begin{verbatim}
The provided conversations (JSON structures) in Polish
are transcribed and divided into turns. A 'turn' is
the continuous utterance of a speaker participating
in a dialogue with at least one other person.

Besides the core grammatically coherent structure of
an utterance, its transcription may include disruptions
or extra elements commonly found in spoken language:
- pauses: '...', (...) and '(yy)'
- repetitions, substitutions and reformulations
- restarts

Remove these speech-specific disruptions and extra
elements from the input turn and output the JSON
structure with a list of cleaned-up turns:

INPUT:
```json
{
    cbiz_tc_53: [
     "(...) Dzień... dzień dobry pani.",
        "(yy) Czy ma pan przygotowany (yy) kod siedmio... 
         (yy) ośmiocyfrowy?",
     "W związku z sytu... z obecną sytuacją",
        "To teraz część ... to ja pana teraz przekierowuję
         do części automatycznej."
    ]}
```

OUTPUT:
```json
{
    cbiz_tc_53: [
     "dzień dobry pani.",
     "Czy ma pan przygotowany kod ośmiocyfrowy?",
     "W związku z obecną sytuacją",
     "to ja pana teraz przekierowuję do części
         automatycznej."
    ]}
```

Explanation of the above example:
- 1. turn: pauses and repetition are removed
- 2. turn: pauses and substitutions are removed
- 3. turn: pause and reformulation are removed
- 4. turn: pause and restart are remove

Keep the grammatically correct and coherent parts of
the turn. Note that a list or a non-verbal sentence
is considered an acceptable utterance.

DO NOT insert additional words or characters.
DO NOT modify input words.
The input and output transcriptions MUST have
the same number of turns.
\end{verbatim}
\caption{JSON-based prompt in English.}
\label{fig:json_prompt}
\end{figure}

\clearpage
\section{Appendix}
\label{appendixC}

We analyse the out-of-vocabulary words newly generated by LLMs in detail and categorise them into (LLMs' outputs are highlited in green):
\begin{enumerate}
\item \textbf{corrections} of grammatical errors and typos:\\
$\bullet$ {
\textsl{zadzwonię}$_{[\textsc{future tense}]}$ (Eng. \textit{I will call)} $\rightarrow$ \textcolor{greeny}{zadzwoniłem}$_{[\textsc{past tense}]}$ (Eng. \textit{I called})}\\
$\bullet$ {
\textsl{płatności}$_{[\textsc{singular number}]}$ (Eng. \textit{payments}) $\rightarrow$ \textcolor{greeny}{płatność}$_{[\textsc{plural number}]}$ (Eng. \textit{a payment})} 
\item  \textbf{completions} of elided words:\\
$\bullet$ {
\textsl{dwóch roboczych} (Eng. lit. \textit{two working}) $\rightarrow$
\textcolor{greeny}{dwóch dni roboczych} (Eng. two working days)}
\item \textbf{questionable morphological modifications}:\\
$\bullet$ {
aspect change: \textsl{nastawiałabym się}$_{[\textsc{imperfective}]}$ (Eng. \textit{I would set myself up}) $\rightarrow$ \textcolor{greeny}{nastwiłabym się}$_{[\textsc{perfective}]}$}\\
$\bullet$ {
gender change: \textsl{zajęłam}$_{[\textsc{feminine}]}$ (Eng. I occupied) $\rightarrow$ \textcolor{greeny}{zająłem}$_{[\textsc{masculine}]}$}
\item \textbf{completing false starts} instead of removal:\\ 
$\bullet$ {
\textsl{\underline{Rozumiem, że... (yy) jeszcze raz jakbym...} Przepraszam, jakby mogła pani jeszcze powtórzyć} (Eng. I understand that... (yy) again I'm like... I'm sorry, could you repeat once again) $\rightarrow$ 
\textcolor{greeny}{\underline{Rozumiem, że $[$chodzi o płatność kartą$]$}. Przepraszam, jakby mogła pani jeszcze powtórzyć}}.
\item Adding \textbf{English translations} instead of or with Polish output:\\
$\bullet$ {
\textit{(yy) Tak, potwierdzam.} (Eng. \textit{(yy) Yes, I confirm.}) $\rightarrow$ \textcolor{greeny}{Tak, potwierdzam. (I confirm.)}}
\item Adding \textbf{explanations}:\\
$\bullet$ {
\textit{Aha.} $\rightarrow$ \textcolor{greeny}{Aha. (This is a non-verbal sound and not considered an utterance.)}}
\item \textbf{Incorrect language identification}:\\
$\bullet$ {
\textit{No SMS-em} (Eng. \textit{Well, by text message})  $\rightarrow$ \textcolor{greeny}{Brak SMS-ów} (Eng. \textit{No SMS-s}).}
\end{enumerate}

\end{document}